\documentclass[times,referee,twocolumn,final,authoryear]{elsarticle}

\usepackage{framed,multirow}
\usepackage{lineno,hyperref}
\usepackage{amssymb,amsmath}
\usepackage{latexsym}
\usepackage{url}
\usepackage{amsmath}
\usepackage{algorithm}
\usepackage[noend]{algpseudocode}
\makeatletter
\def\BState{\State\hskip-\ALG@thistlm}
\makeatother

\usepackage{url}
\usepackage{xcolor}
\usepackage{multirow,color}
\definecolor{snm}{rgb}{.8,.349,.1}
\newcommand{\snm}[1]{\textcolor{snm}{#1}}

\journal{Computer Vision and Image Understanding}

\begin{document}














\begin{frontmatter}

\title{Informative sample generation using class aware generative adversarial networks for classification of chest Xrays}

 \author[1,11]{Behzad \snm{Bozorgtabar}} 

 \author[2]{Dwarikanath \snm{Mahapatra} \corref{cor1}}
 \cortext[cor1]{Corresponding author: 
   Tel.: +61-424-128-296;  
 }
 	\ead{dmahapatra@gmail.com}
	
 \author[3]{Hendrik \snm{von Teng}}
 \author[3]{Alexander \snm{Pollinger}}
 \author[3]{Lukas \snm{Ebner}}
 \author[1,4]{Jean-Phillipe \snm{Thiran}}
 \author[5]{Mauricio \snm{Reyes}}

 \address[1]{Signal Processing Laboratory (LT55), Ecole Polytechnique 
Féderale de Lausanne (EPFL-STI-IEL-LT55), Station 11, 
1015 Lausanne, Switzerland}
\address[11]{Center for Biomedical Imaging, Lausanne, Switzerland}
 \address[2]{IBM Research Australia, Melbourne 3006, Australia}
 \address[3]{University Institute for Diagnostic, Interventional, and Pediatric Radiology, University Hospital Bern, Switzerland}
 \address[4]{Department of Radiology, University Hospital Center 
(CHUV), University of Lausanne (UNIL), Lausanne, Switzerland}
 
 \address[5]{University of Bern, Bern 3014, Switzerland} 


\begin{abstract}
Training robust deep learning (DL) systems for disease detection from medical images is challenging due to limited images covering different disease types and severity. The problem is especially acute, where there is a severe class imbalance. We propose an active learning (AL) framework to select most informative samples for training our model using a Bayesian neural network. Informative samples are then used within a novel class aware generative adversarial network (CAGAN) to generate realistic chest xray images for data augmentation by transferring characteristics from one class label to another. Experiments show our proposed AL framework is able to achieve state-of-the-art performance by using about $35\%$ of the full dataset, thus saving significant time and effort over conventional methods.
\end{abstract}


\end{frontmatter}



\section{Introduction}
\label{sec:intro}

Medical image classification  is an essential component of computer aided diagnosis systems, where deep learning (DL) approaches have led to state-of-the-art performance \citep{DLRevTMI}. Robust DL approaches need large labeled datasets, which is challenging to obtain for medical images because of: 1) limited expert availability; 2) intensive manual effort required to curate (i.e. label) datasets; and 3) paucity of images for specific disease labels leading to class imbalance.
 In most public medical image databases, there is a severe class imbalance. For example in the NIH chest Xray-14 dataset \citep{NIHXray} more than $50\%$ of the patient samples consist of normal cases, and the remaining dataset has samples from $14$ different disease types. The diseased samples are not uniformly distributed with the least represented case comprising less than $2\%$ of the diseased samples while the most represented case has more than $20\%$ of the diseased samples. 

Consequently, algorithms trained on this dataset will be biased towards the majority class. Although a weighted cross entropy cost function may be used to overcome this imbalance, in practice we find that this is not optimal for datasets with multiple classes (in this case $14$). To overcome the limitations of unbalanced datasets we propose an active learning (AL) based approach to identify informative samples and also generate more samples for the minority classes. Our approach has the advantage of: 1) training the algorithm on the most informative samples so as to achieve at-par performance with the state-of-the-art methods, albeit with much less labeled data; 2) ensuring equal representation of all classes to improve the robustness of the trained classifier.

Existing active learning approaches overcome data scarcity by incrementally selecting the most informative unlabeled samples, querying their labels and adding them to the labeled training set \citep{AL1}. AL systems are first initialized with a relatively small set of labeled data. In subsequent iterations, the most informative samples (and corresponding labels) are queried and added to the training set to update the model. 
%
AL in a DL framework poses the following challenges: 1) labeled samples generated by current AL approaches are too few to train or fine-tune convolutional neural networks (CNNs); 2) Existing AL methods select informative samples using application specific metrics, while feature learning and model training are jointly optimized in CNNs.  Thus fine-tuning CNNs in the traditional AL framework may not lead to model convergence \citep{AL_CEAL}. We propose a AL based framework using Class Aware Generative Adversarial Networks (CAGANs)  for synthetic sample generation and use it for CNN based classification of medical images. We achieve state-of-the-art results using fewer real images while we benefit from synthetic images.

\subsection{Related Work}
GANs \citep{GANs} have been used in various applications such as image super resolution \citep{SRGAN,MahapatraMICCAI_ISR}, image synthesis and image translation using conditional GANs (cGANs) \citep{CondGANS} and cycle GANs (cycleGANs) \citep{CyclicGANs}. Frid-Adar et al. \citep{AdarISBI18} propose a data augmentation method using GANs that synthetically enlarges the training dataset for model improvement. In the standard GAN framework \citep{GANs}, the generator learn a mapping from random noise vector \emph{z} to output image \emph{y}: $G : z \rightarrow y$. Conditional GANs (cGANs) \citep{CondGANS} learn a mapping from observed image \emph{x} and random noise vector \emph{z}, to \emph{y}: $G :\{x,z\} \rightarrow y$. 

Semi-supervised learning (SSL) \citep{Chapelle} and
  active learning \citep{SettlesCraven2008} methods have been used to overcome the limitations of insufficient labeled samples in medical applications such as segmenting anatomical structures \citep{IglesiasActive,MahapatraMICCAI_CD2013} and detecting cancerous regions \citep{DoyleBMC11}. An important component of AL systems is the query strategy to select the most ``informative'' sample and popular approaches include uncertainty sampling \citep{Lewis94}, query-by-committee (QBC) \citep{Freund97} and density weighting \citep{SettlesCraven2008}. An excellent review of different query strategies is given in \citep{SettlesCraven2008}. 
  
  In recent work on using AL with deep neural networks (DNNs), Gal et. al. \citep{Gal2017Active} use Bayesian deep neural networks to design an active learning framework for high dimensional data. Wang et al. \citep{AL_CEAL} propose a `cost-effective AL approach for computer vision applications, which leverages unlabeled data with high classification uncertainty and high confidence to update the classifier. Yang et al. \citep{YangAL_MICCAI17} use AL with fully convolutional networks (FCN)  for segmenting histopathology images.
  
  In our recent work \citep{MahapatraMICCAI2018}, we proposed a conditional GAN (cGAN) based approach for AL based medical image classification and segmentation. 
  As our second contribution, we hypothesized that a time-effective (i.e. reduced expert monitoring/correction) active learning approach  can be obtained by joining the process of sample selection, via its informativeness, and data augmentation for model training improvement. In particular, we proposed image synthesis model that learns to generate realistic images from samples chosen by means of model uncertainty using a Bayesian deep neural network. Thus, the system not only generates realistic samples, but utilizes the model uncertainty estimates on input samples to drive its data augmentation process on challenging samples. Promising preliminary results were obtained for image segmentation and classification of lung images. 
 
In this paper we build on these preliminary results of \citep{MahapatraMICCAI2018} and expand the study to a much larger dataset. In addition, we perform a deeper comparison to standard data augmentation approaches and analyze the benefits of utilizing a proposed GAN based model, by comparing it to a regular GAN for synthetic data generation. Finally, we extend the methodology to a Class-Aware Generative Adversarial Network (CAGAN) to incorporate a class balancing component, which is able to transfer image characteristics across samples of different classes.

We test the proposed approach for the key task of medical image classification for disease detection, demonstrating its ability to yield models with high accuracy while reducing the number of samples required for training. 
The rest of the paper is arranged as follows: Section~\ref{sec:met} describes our method and highlights our active learning framework. Section~\ref{sec:expts} describes the dataset, the detailed implementation of our method and the results of different experiments. Finally we conclude with Section~\ref{sec:concl} with a summary of our findings and potential future work.

\subsection{Contributions}

Our paper makes the following contributions: 
\begin{enumerate}
\item  We propose a Class-Aware Generative Adversarial Network based AL framework that boosts a classifier using synthetically generated samples and transfers learnt characteristics between different classes; 
\item  We show that use of our approach leads to realistic generated images that contribute to improving the performance of a classification system despite having very small number of real training samples. Further, compared to GANs, we show that a better performance can be attained through a proposed CAGAN model to focus the data augmentation process on samples the model has difficulties with.
\item The synthetically generated images are able to alleviate the common problem of class imbalance in medical image analysis tasks thus resulting in more robust classification systems.
\end{enumerate}


\section{Methods}
\label{sec:met}
We first describe our approach (CAGAN) in \ref{met:cGAN}, then the proposed AL approach will be discussed in \ref{met:uncertainty}. The former generates realistic samples that introduces meaningful information in the training mode, while the latter identifies informative samples to improve model performance. This is different from conventional AL approaches, which identify informative samples using an uncertainty measure. Since medical datasets acquired in a few locations under similar conditions offer limited diversity, uncertainty may not be very effective due to model bias. Our framework is able to overcome this limitation using three components for: 1) sample generation; 2) classification model; and 3) sample informativeness calculation.


To show the usefulness of the generated images, we employ a data augmentation strategy for our experiments. A few annotated samples are used to finetune a pre-trained VGG16 \citep{VGG} (or any other classification network) using standard data augmentation such as rotation, translation and scaling. The sample generator takes an input image of any class (the base class label) and generates realistic looking images of the specified class label which may be different from the base label. A Bayesian neural network (BNN) \citep{BDNN} calculates generated images' informativeness and highly informative samples are added to the labeled image set. 
The new training images are used to fine-tune the previously trained classifier. This updated classifier is tested on a separate validation set  which is different from the training and test set and whose size does not change. If the classification performance of the updated classifier on the validation set (in terms of AUC values)  does not change significantly for such updates then we assume there are no more informative images and we stop the sample selection. 

The training steps are summarized in Algorithm \ref{alg1}. Note that we split our entire labeled set into training, test and validation folds. Network tuning is done using the training and validation set while all reported performance metrics are on the test set.
	
\subsection{Class Aware Generative Adversarial Network}
\label{met:cGAN}
We present a variant of the GAN model namely Class Aware Generative Adversarial Network (CAGAN). Here, our objective is to train a generator $G$ that translates an input image $x$ into an output image $y$ conditioned on the latent vector $z$ (obtained from a pre-trained autoencoder that takes as input the output of a segmentation network), i.e., $G(x,z) \rightarrow y$. Additionally, we introduce class information to a GAN by adding an explicit class loss term that allows the discriminator to produce probability distributions over both sources and class labels. It should be noted that the proposed approach is different from conditional GAN, where image attributes (e.g. classification categories) are fed into both generator and discriminator, respectively.

The segmentation network mentioned earlier is a UNet \cite{UNet} that is trained to segment both lungs from an input xray image. This is done to get an automated segmentation output. During training phase, we use the manual segmentation masks of available images and this network can be used for new images that may be added to the training set. The UNet architecture consists of an encoder part with 4 convolutional layers each with filter size of $3\times3$ and $64$ filters in each layer. The output of the final convolution layer is connected to a $256$ dimensional fully connected layer. This is then connected to the decoder path with $4$ convolution layers having $64$ filters of size $3\times3$ in each layer.  

\begin{algorithm}{}
\label{alg1}
\begin{algorithmic}[1]
\Procedure{Focused Active learning based classification}{}
\State U-Net $\gets$ \textit{Pre-trained Segmentation Network }
\State CN $\gets$ \textit{Initialize Classification Network }
\State CAGAN $\gets$ \textit{Initialize Class Aware Generator Network }
\BState \emph{loop}:
\State For each input image $I$ and known target class $c$
\State Generate latent code $z$ from U-Net
\State Use CAGAN to generate synthetic images of specified class $c$,  using $z$ and input image $I$. $\{\hat{I},c\}=G\left ( I,z \right)$
\State Quantify informativeness of generated images $\hat{I}$ using Eq. \ref{eqn:Uncert}.
\State Sort generated images in descending order of informativeness and select top $N_{Inf}$ informative ones
\State Re-train CN with new added $N_{Inf}$ informative images
\State Check performance on separate validation set
\State If AUC on validation set improves, continue to \emph{loop}.
\State \textbf{close}
\State \textbf{Outputs}: Trained CN and CAGAN
\EndProcedure
\end{algorithmic}
\end{algorithm}

\subsubsection{Adversarial Loss}
To synthesize realistic images, we adopt an adversarial loss $L_{adv}$. This reassures our network to favor solutions that reside on the manifold of natural images, by seeking to fool the discriminator network. 

\begin{equation}
L_{adv} =\mathbb{E}_x [\log D_{src}(x)] + \mathbb{E}_{x,z}[\log (1 - D_{src}(G(x,z))],
\label{eq:sgan1}
\end{equation}
where $G$ generates a fake image $x_{fake}=G\left ( x,z \right )$ conditioned on both an observed image $x$ and the latent vector $z$, while $D$ seeks to detect which examples are fake. The term $D_{src}(x)$ denotes a probability distribution over sources given by $D$. 

\subsubsection{Classification Loss}
Now, let assume that we are given the input image and classification label pair $(x,c)$ as part of the training data. For a given input image $x$ and a target label $c$, our goal is to generate output image, which is properly classified to the corresponding class label $c$. To do this, we add an auxiliary classifier on top of $D$ and impose the classification loss when optimizing both networks $D$ and $G$. We disentangle the classification loss into two terms: a classification loss of real images $L^{r}_{cls}$ used to optimize $D$, and a classification loss of fake images $L^{f}_{cls}$ used to optimize $G$. The classification loss of real images is defined as:

\begin{equation}
L^{r}_{cls} = \mathbb{E}_{x,c} [-\log D_{cls}(c|x)],
\label{eq:sgan2}
\end{equation}
where the term $D_{cls}(c|x)$ represents a probability distribution over classification labels computed by $D$. The class loss term indicates to the model that the information we classify on will generate images, which match the target labels $c$. On the other hand, the loss function for the classification of fake images is defined as:
\begin{equation}
L^{f}_{cls} =\mathbb{E}_{x,z,c}[- \log D_{cls}(c|G(x,z))].
\label{eq:sgan3}
\end{equation}

In above equation, $G$ is optimized to minimize this objective for correct prediction on the classifier, which demonstrates results closer to the original ones.

\subsubsection{Content Loss}
Using the adversarial and classification losses, the generator is trained to generate realistic images with the corresponding classes. However, it does not guarantee that synthetic images preserve the content of input images. To address this problem , we use content loss ($L_{content}$):  
\begin{equation}
L_{content} =  VGG(x,y) + MSE(x,y) + \frac{1}{NMI(x,y)+0.0001}
\label{eq:conLoss}
\end{equation} 
$NMI$ denotes the normalized mutual information (NMI) between $x$ and $y$, and is used to determine similarity of image's intensity distribution. $MSE$ is the intensity mean square error. In addition, we use the VGG loss based on the ReLU activation layers of the pre-trained VGG-16 network described in \citep{VGG}. We then define the VGG loss as the $l2$ distance between the feature representations of a reconstructed image and real image using Relu 4-1 layer. For similar images, $NMI$ gives higher value while $VGG$ and $MSE$ give lower values. 

\subsubsection{Overall Objective} 
Finally, the generator $G$ and discriminator $D$ are trained with a linear combination of the loss terms:

\begin{equation}
L_D = -L_{adv} + \lambda_{cls} L^{r}_{cls}
\end{equation}
\begin{equation}
L_G = L_{adv} + \lambda_{cls} L^{f}_{cls}+\lambda _{content}L_{content} 
\end{equation}
where $\lambda_{cls}$ and $\lambda_{content}$ are hyper-parameter that control the relative importance of classification loss and content loss, compared to the adversarial loss. We set $\lambda_{cls} = 1$ and $\lambda_{content} = 10$ in all of our experiments.

\subsubsection{Importance of latent space information} 
As explained earlier we use the latent vector encoding of the lung mask in the CAGAN to generate realistic xray images. The primary motivation in doing so is to condition the image generation on the lung shape and add additional  information to make the generated images more realistic. 

Another factor behind the use of latent shape information is to change the shape of the lung masks and generate a diverse set of images with different image characteristics.
Change in lung shape is effected by changing the contour of the lung mask and using b-splines to get a smooth shape.

%

\subsection{CAGAN Implementation}
It is non-trivial to train GAN models in a stable manner. To address this problem, we generalize the state-of-the-art Wasserstein GAN to the context of image generation for more stable convergence. In fact, we replace Eq.~\ref{eq:sgan1} with a Wasserstein GAN having gradient penalty defined as:

\begin{equation}
\begin{split}
L_{adv} =& \mathbb{E}_x [D_{src}(x)] - \mathbb{E}_{x,z}[D_{src}(G(x,z))] \\ 
& - \lambda_{gp} \mathbb{E}_{\widehat{x}}[(\left\|\nabla_{\widehat{x}} D_{src}(\widehat{x})\right\|_2 - 1)^{2}] 
\end{split}
\end{equation}

where $\widehat{x}$ is sampled uniformly along a straight line between
a pair of a real and a generated images and $\lambda_{gp}$ is set to 10 in all experiments.

\subsection{CAGAN Architecture}
CAGAN has the generator network composed of three convolution layers with a stride size of two, followed by batch normalization and ReLU activation for downsampling, three residual blocks, and three transposed convolution layers with the stride size of two for upsampling before the last convolutional layer. We utilize PatchGANs \citep{li2016precomputed} for the discriminator network, which classifies whether local image patches are real or fake.

\begin{figure*}[t]
\centering
\includegraphics[height=10.5cm, width=15.8cm]{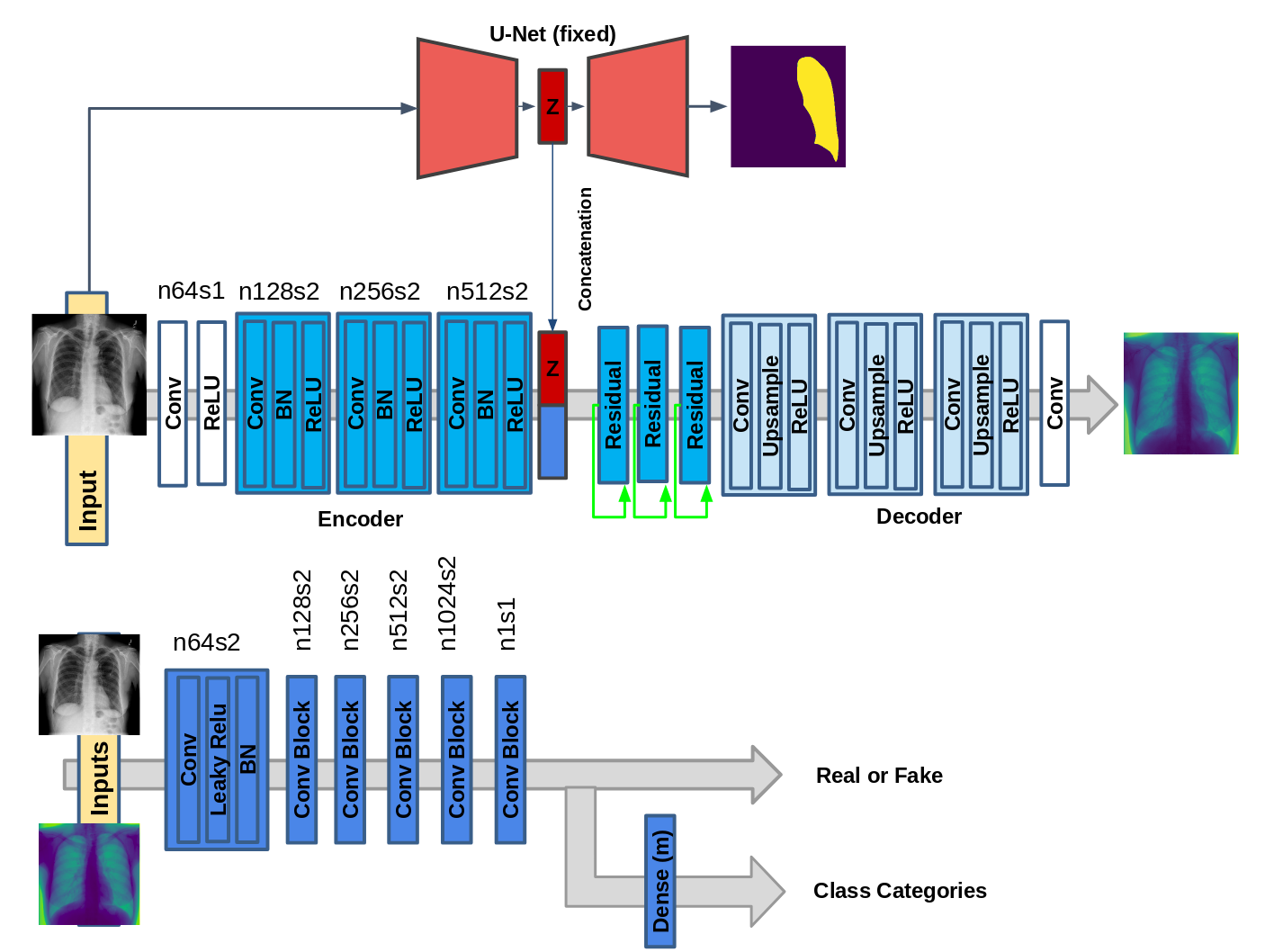}
  \caption{Architecture of generator network (\textbf{top}) and discriminator network (\textbf{bottom}). Latent code $z$ is injected from pre-trained U-Net by spatial replication and concatenation into the generator network and $m$ is the number of classes.}
\label{fig:Gan}
\end{figure*}

The discriminator $D$ has six convolution layers with the kernels increasing by a factor of $2$ from $64$ to $1024$. Leaky ReLU is used and strided convolutions reduce the image dimension when the number of features is doubled. The last convolutional layer is followed by dense layer to output probabilities for different classes. See Figure~\ref{fig:Gan} for schematic diagrams of CAGAN. 

\subsubsection{Synthetic Image Generation}
During the test stage, the generator takes as input the test Xray image and the latent vector encoding of a mask (either original or altered) and outputs a realistic Xray image whose label class is the same as the original image.   
%

\begin{figure}[h]
\begin{tabular}{ccc}
\includegraphics[height=2.9cm, width=2.3cm]{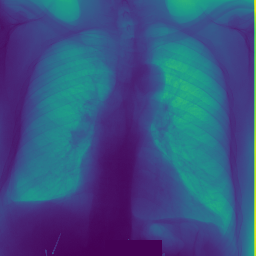}  &
\includegraphics[height=2.9cm, width=2.3cm]{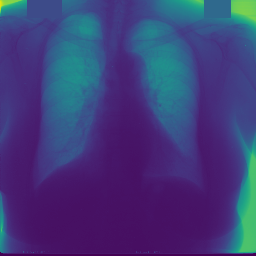}  &  
\includegraphics[height=2.9cm, width=2.3cm]{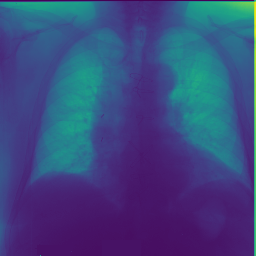}   \\
(a) & (b) & (c)  \\
\includegraphics[height=2.9cm, width=2.3cm]{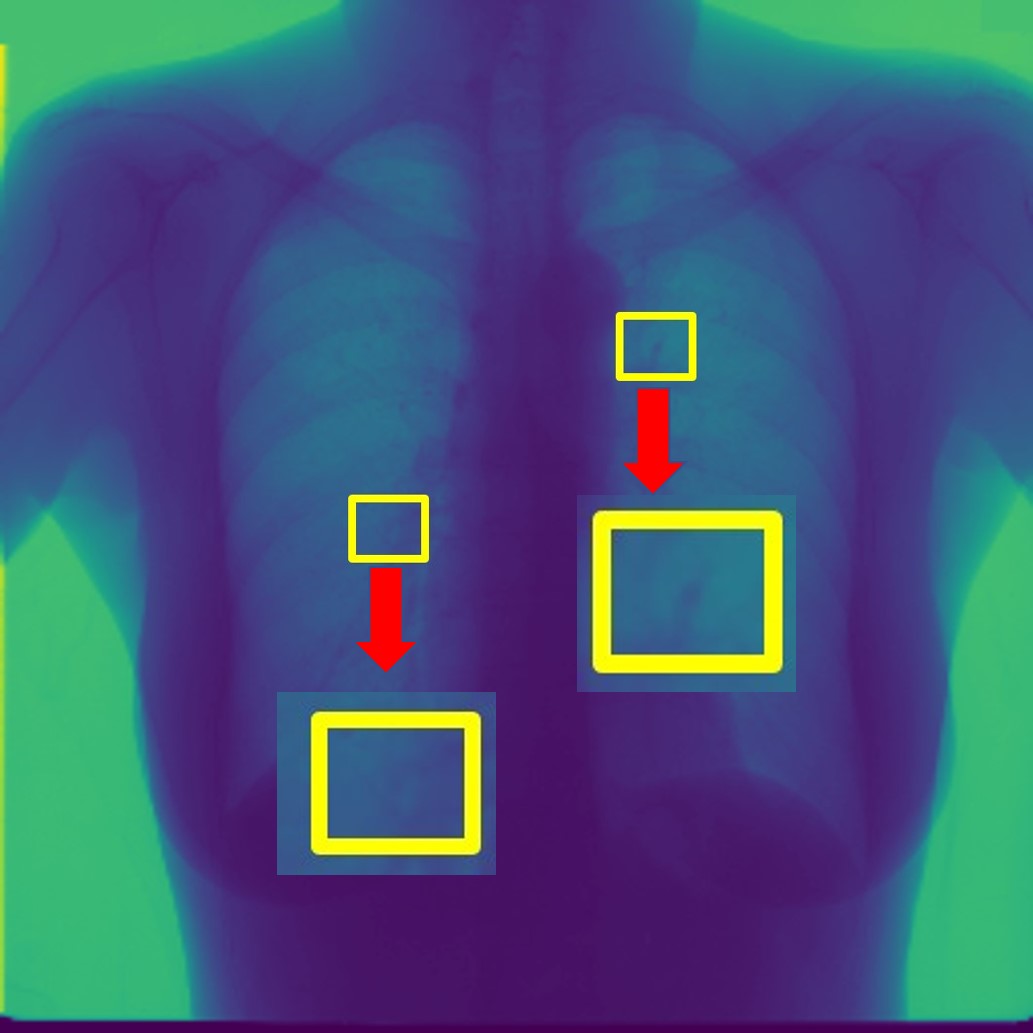}  &
\includegraphics[height=2.9cm, width=2.3cm]{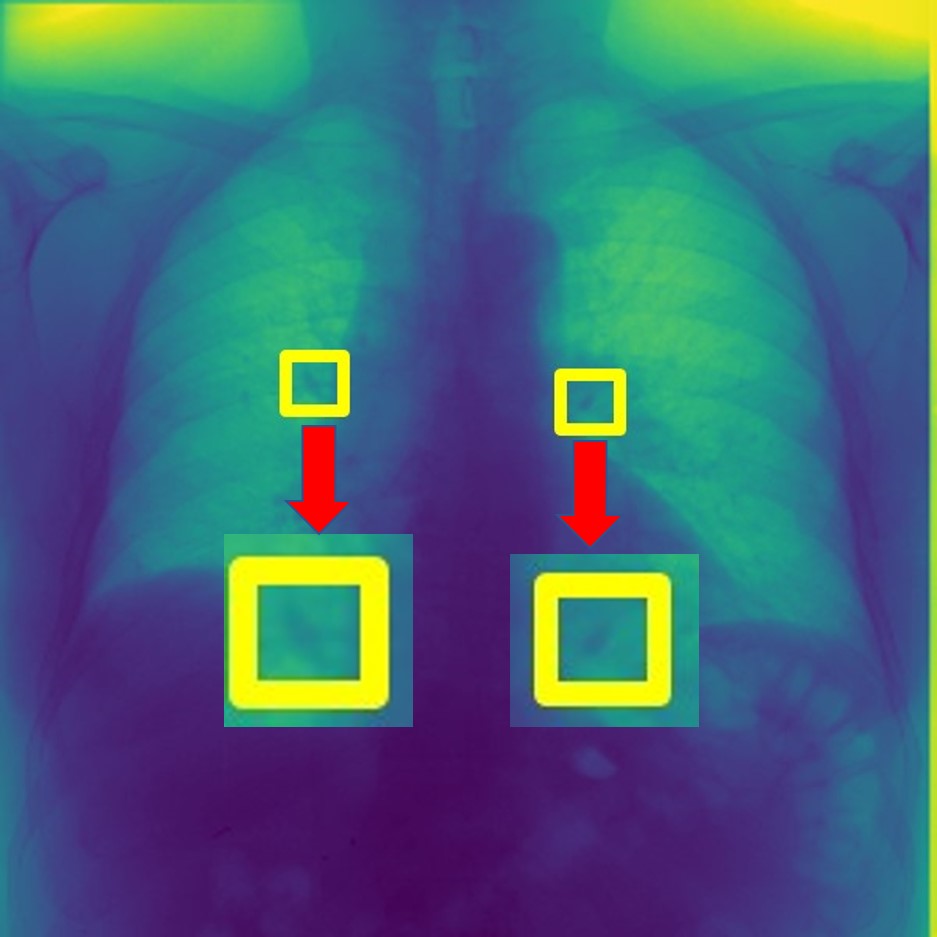}  &
\includegraphics[height=2.9cm, width=2.3cm]{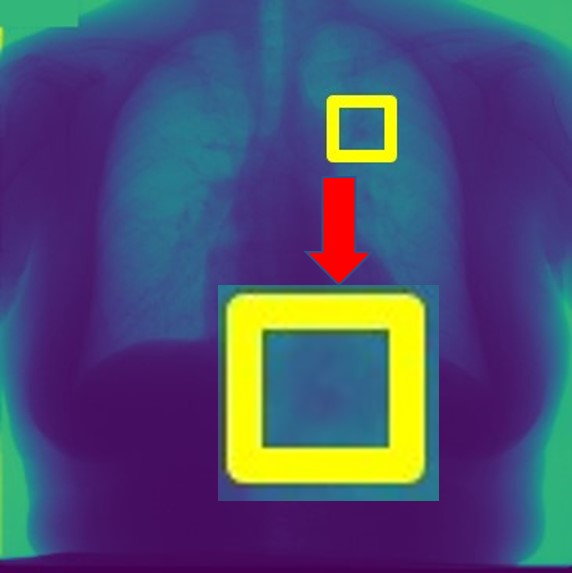}  \\
(d) & (e) & (f)  \\
\end{tabular}
\caption{Generated informative xray image; (a)-(c) non-diseased cases; (d)-(f) images with nodules of different severity at the center of yellow box. (a), (d) are the original images while others are generated synthetic images.}
\label{fig:GenImages}
\end{figure}

\begin{figure}[h]
\begin{tabular}{ccc}
\includegraphics[height=2.4cm, width=2.3cm]{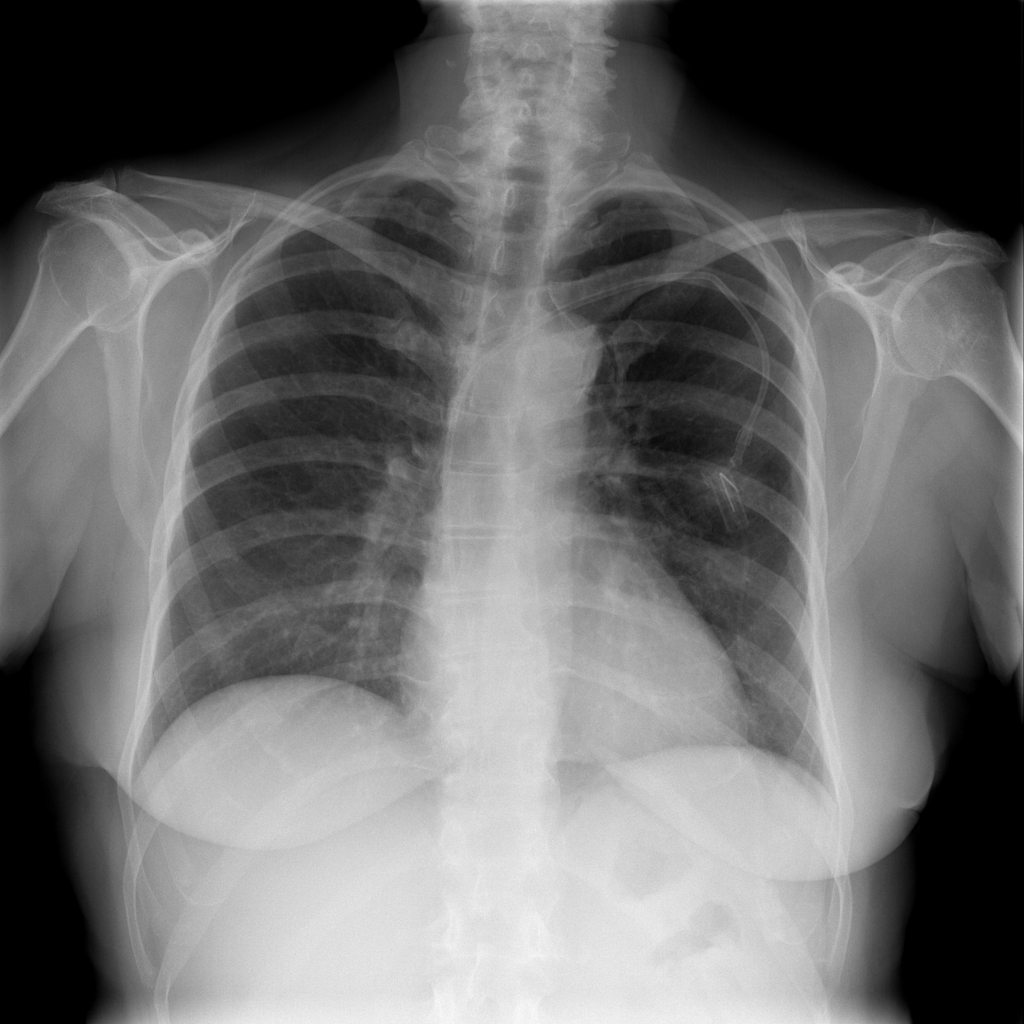}  &
\includegraphics[height=2.4cm, width=2.3cm]{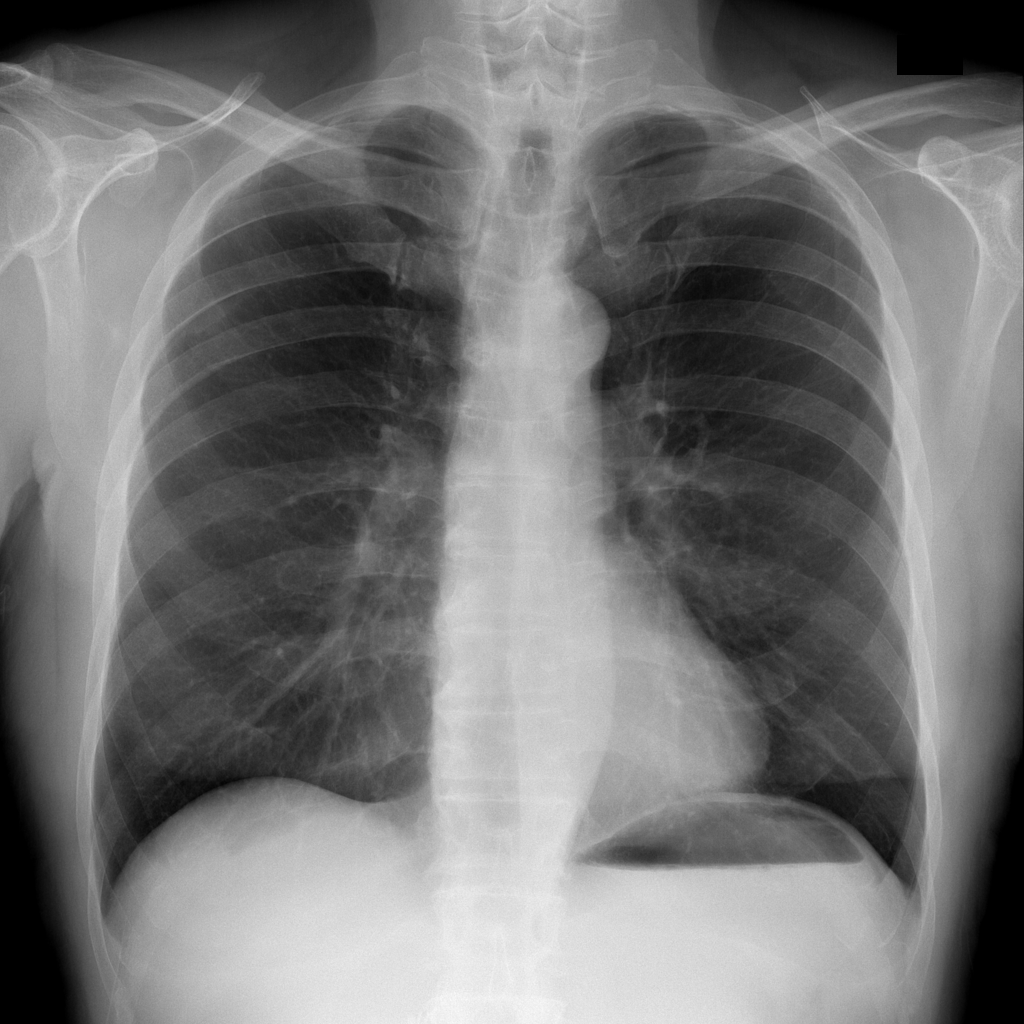}  &  
\includegraphics[height=2.4cm, width=2.3cm]{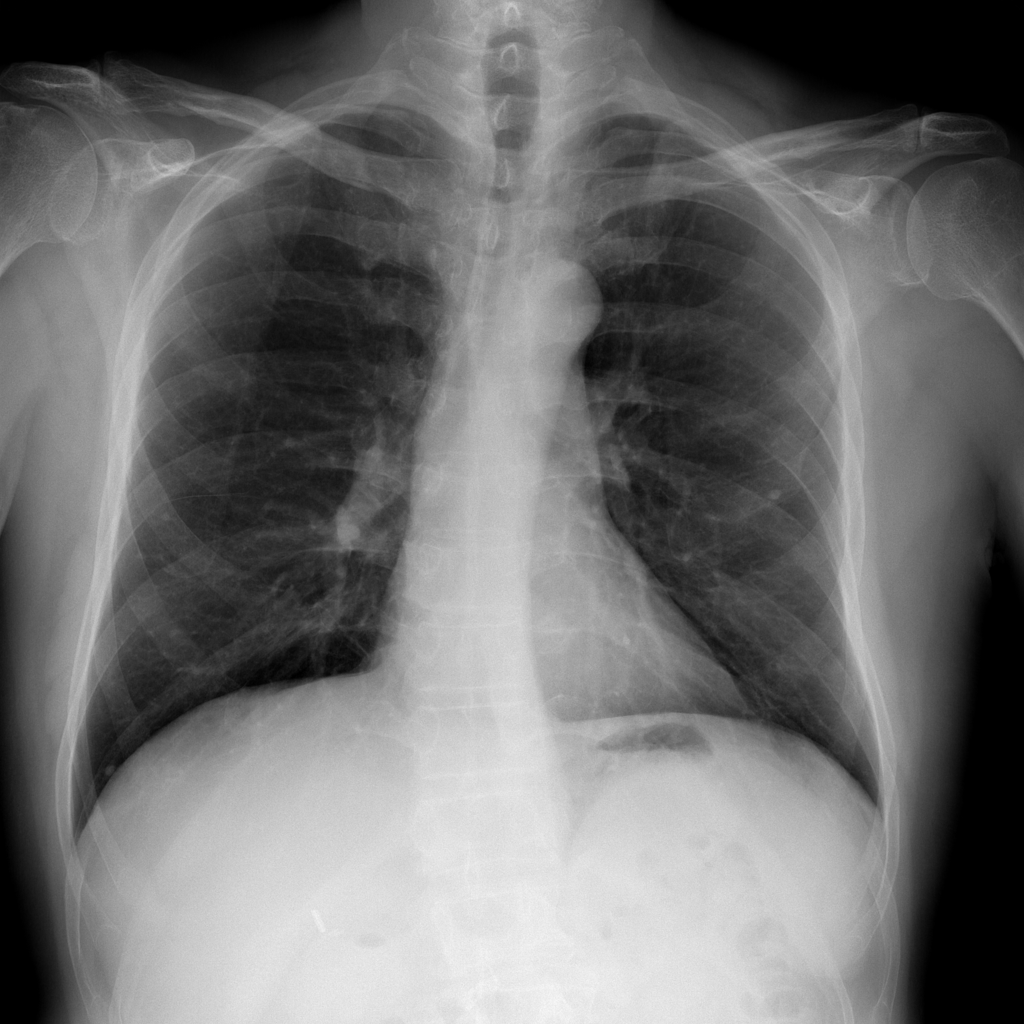} \\ 
\includegraphics[height=2.4cm, width=2.3cm]{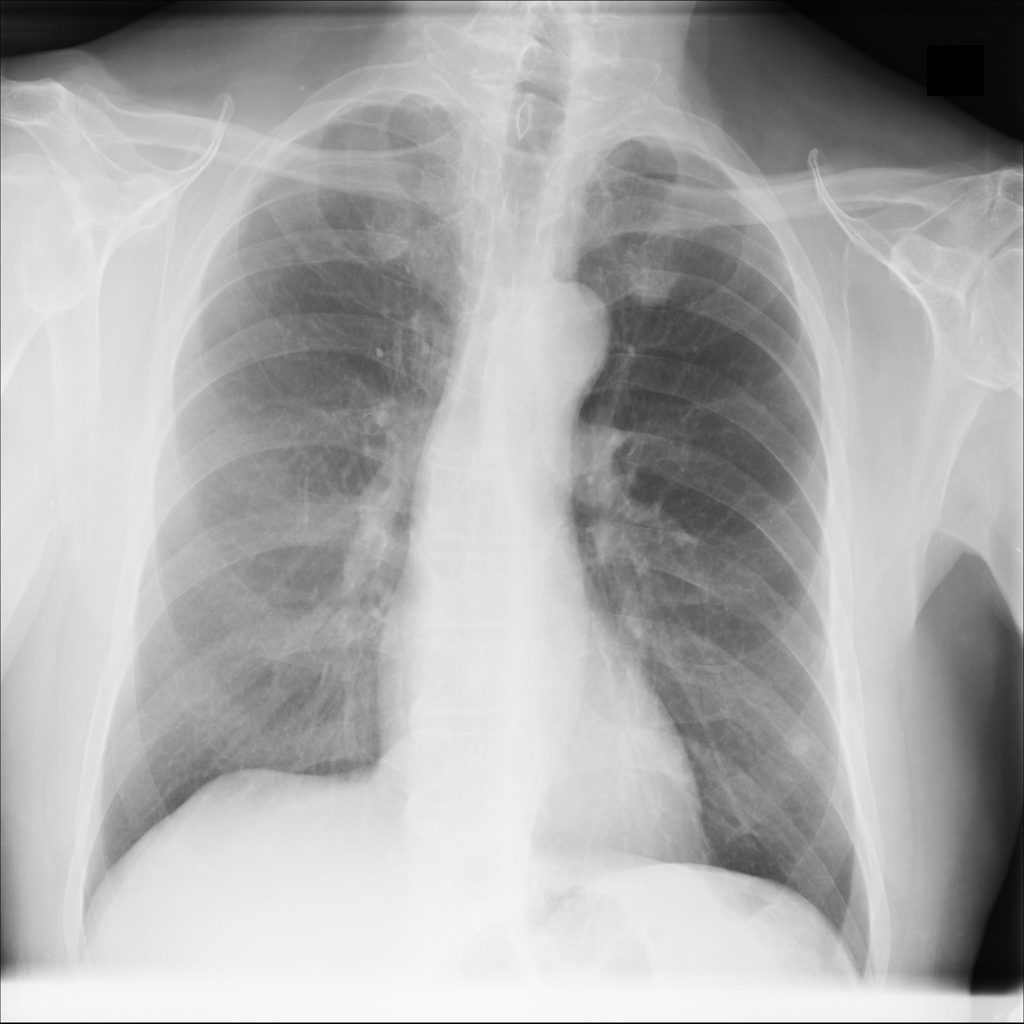}  &
\includegraphics[height=2.4cm, width=2.3cm]{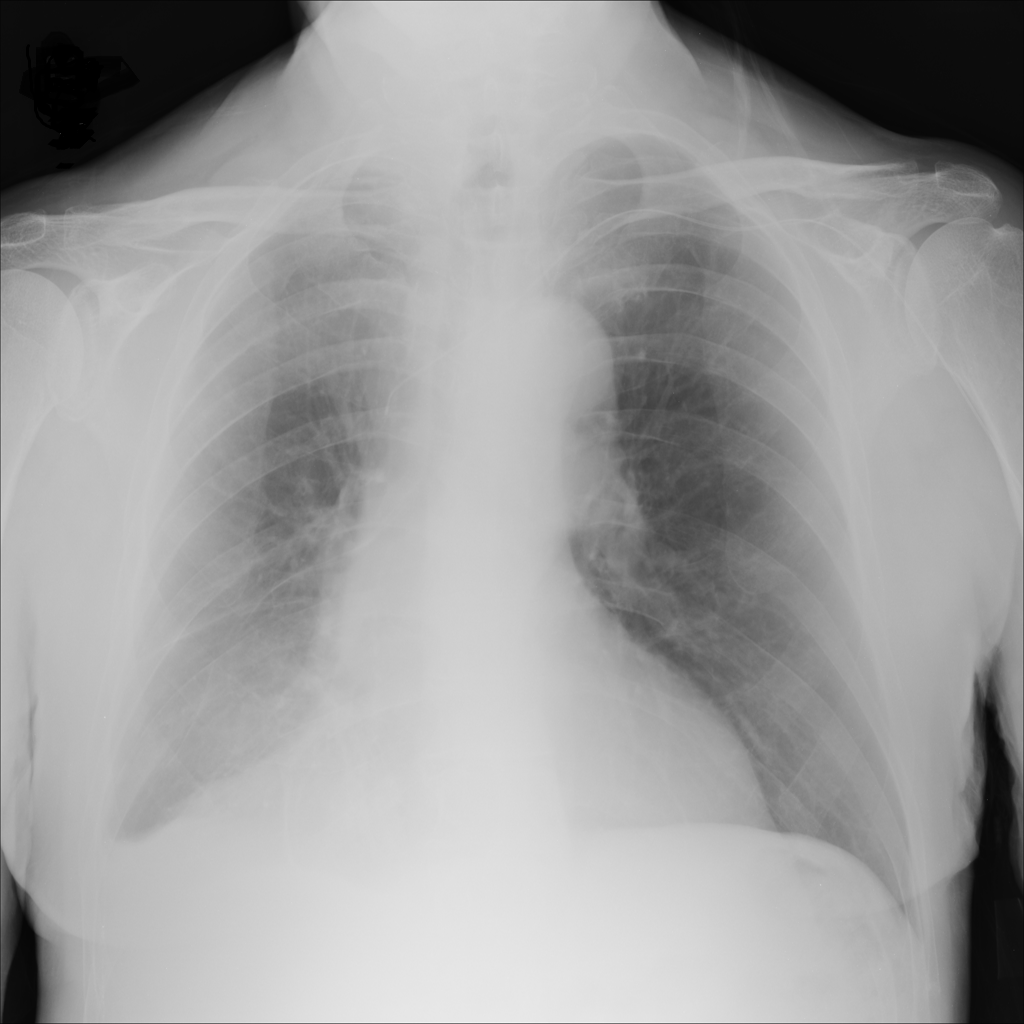}  &
\includegraphics[height=2.4cm, width=2.3cm]{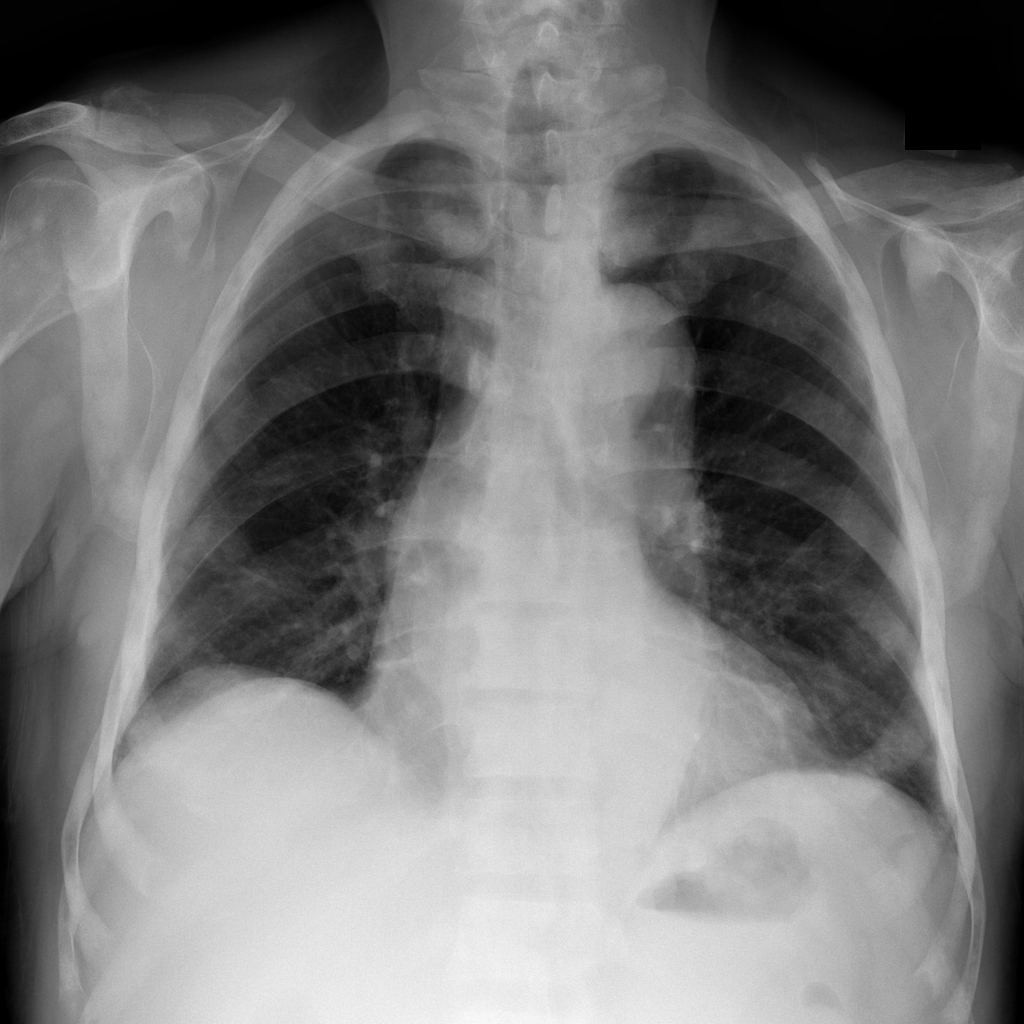} \\ 
\end{tabular}
\caption{Real and generated informative xray image. Top row shows the real images from the NIH Xray 14 dataset \citep{NIHXray} and the bottom row shows generated images using our method.}
\label{fig:GenImages1}
\end{figure}

For every test image, we obtain up to $200$ modified masks (by changing the lung mask) and their generated images. 
Figure~\ref{fig:GenImages1} shows examples of real and generated images from the NIH dataset \citep{NIHXray}. The top row shows the real images while the bottom row shows images obtained using our method.
Figure~\ref{fig:GenImages} (a) shows an original normal image from the SCR chest Xray dataset \citep{SCR} while Figs.~\ref{fig:GenImages} (b,c) show generated `normal' images similar to real dataset. Figure~\ref{fig:GenImages} (d) shows an image with nodules from the SCR chest Xray dataset, and Figs.~\ref{fig:GenImages} (e,f) show generated `nodule' images. Although the nodules are very difficult to observe with the naked eye, we highlight its position using yellow boxes. It is quite obvious that the generated images are realistic and suitable for training. Other examples of real and generated images from the NIH database are shown in Figure~\ref{fig:GenImages1}.

\subsection{Sample Informativeness Using Uncertainty}
\label{met:uncertainty}

Each generated image's uncertainty is calculated using the method described in \citep{BDNN}. In short, two types of uncertainty measures can be calculated from a Bayesian neural network - epistemic and aleotaric uncertainty. Aleotaric uncertainty models the noise in the observation while epistemic uncertainty models the uncertainty of model parameters, i.e., how uncertain is the model in correctly classifying the image sample. We adopted \citep{BDNN} to calculate uncertainty by combining the above two types. We give a brief description below and refer the reader to \citep{BDNN} for details.
For a BNN model $f$ mapping an input image $x$, to a unary output $\widehat{y}\in R$, the predictive uncertainty for pixel $y$ is approximated using:
\begin{equation}
Var(y)\approx \frac{1}{T} \sum_{t=1}^{T} \widehat{y}_t^{2} - \left(\frac{1}{T} \sum_{t=1}^{T} \widehat{y}_t \right)^{2} + \frac{1}{T} \sum_{t=1}^{T} \widehat{\sigma}_t^{2}
\label{eqn:Uncert}
\end{equation}
$\widehat{\sigma}^{2}_t$ is the BNN output for the predicted variance for pixel $y_t$, and ${\widehat{y}_t,\widehat{\sigma}^{2}_t}^{T}_{t=1}$ being a set of $T$ sampled outputs.

\section{Experiments}
\label{sec:expts}

\subsection{Dataset Description}

We apply our method to the NIH ChestXray14 dataset \citep{NIHXray}, which has $112,120$ frontal-view X-rays from $30,805$ unique patients with $14$ disease labels (multi-labels for each image), which are obtained from associated radiology reports.

The $14$ pathologies are `Cardiomegaly', `Emphysema', `Effusion', `Hernia', `Nodule', `Pneumothorax', `Atelectasis', `Pleural Thickening' (PT), `Mass', `Edema', `Consolidation', `Infiltration', `Fibrosis' and 'Pneumonia'. For further details we refer the reader to \citep{NIHXray}. Table~\ref{tab:AL100} illustrates the percentage distribution of different labels within the \emph{diseased images} only. Note that almost all images have multiple labels associated with them and hence the percentage values do not necessarily add up to $100$. An interesting observation is that `Infiltration' label is shared by $24\%$ of the diseased images while only $0.28\%$ images are labelled `Hernia'. This is indicative of severe class imbalance and our proposed method aims to overcome this common problem in medical image analysis.

\subsection{Implementation Details}
Our method was implemented in TensorFlow. We use Adam \citep{Adam} with $\beta_1=0.93$ and batch normalization. Mean square error (MSE) based ResNet was used to initialize $G$. The final GAN was trained with $10^{5}$ update iterations at learning rate $10^{-3}$. Training and test was performed on a NVIDIA Tesla K$40$ GPU. 

Our experiments focus primarily on classification. Recent work \citep{CheXNet} on the NIH-14 dataset  used a DenseNet architecture \citep{DenseNet} to achieve state  of the art performance in classification. To illustrate the benefits of using our AL approach we explain the working using a $VGG16$ network. We also present results for active learning with other classifiers like ResNet \citep{ResNet} and DenseNet \citep{DenseNet} (which is the basis for CheXNet \citep{CheXNet}). The VGG16, ResNet and DenseNet have all been pre-trained on the Imagenet dataset \citep{Imagenet}.

First, we split our entire dataset into training ($70\%$), validation ($10\%$) and test ($20\%$), at the patient level such that all images from one patient are in a single fold. 
We use a $VGG16$ network \citep{VGG} pre-trained on the Imagenet dataset \citep{Imagenet} as our base network. Initially, we choose samples images from each of the $14$ classes and augment them using rotation and translation. We refer to this step as standard data augmentation (DA). Using this augmented dataset we fine tune the last layer of $VGG16$. 
 From the remaining images in the training dataset the task is to identify the most informative samples from each class and add them to the training set. Recall that different image classes are not equally represented and hence it is essential to  have equal representation of each class in the training batches. 

Using the VGG16 network (after finetuning) we identify the most informative images across all classes and rank them according to their  informativeness  value as calculated using the BNN described in Sec~\ref{met:uncertainty}. Note that the Bayesian network is created by taking the instantaneous $VGG16$ (or any other) classifier after finetuning and modifying its weights. This is done to ensure that the informative samples are determined by the classifier at its current state. We take the top $32$ informative samples irrespective of the class label and use them along with our pre-trained CAGAN network to generate informative samples of other classes.
 
Conventional GANs would be able to generate image samples that have the same class label as the input image. However, as described in Sec~\ref{sec:met}, by using CAGANs we can generate images of different classes despite the input image belonging to a another class. Using the $32$ most informative samples (from the initial pool) we generate $250$ samples of each of the $14$ classes. Subsequently, the $150$ most informative samples of each class are chosen and used to finetune the classifier. The $32$ samples that were chosen as informative are now removed from the labeled training set. The updated classifier is again used to identify informative samples from the updated training set and the entire sequence of steps is repeated. After every update/finetuning the classifier is used to classify the samples in the validation dataset. The iterative sequence of steps (involving informative sample selection, data augmentation and finetuning) continues till the area under curve (AUC) value from the validation dataset does not change by more than $0.1$ for three consecutive iterations. Once this condition is met we consider the classifier as the final model and we use it to determine performance on the test set. 

To compare our AL approach with a fully supervised learning (FSL) based method we use the CheXNet method \citep{CheXNet}. We also finetune $VGG16$ and $ResNet$ classifiers (without sample selection) to obtain baseline performance for the fully supervised case. 
For all the FSL classifiers we use weighted cost functions to account for data imbalance. These set of results serve as the baseline to compare our active learning methods. Table~\ref{tab:AL1} summarizes the results for baseline FSL settings and our proposed method. Note that all performance measures are reported for the test set.

%

\subsection{Classification Results}

In this section, we show results for the active learning (AL) scenario and compare it with conventional fully supervised learning (FSL)  methods that use the same training set for training the model and report results on the test set. The area under curve (AUC) values are reported in each case. For the FSL case we show results for the following classifiers - $VGG16$, $ResNet$ \citep{ResNet} and CheXNet \citep{CheXNet}. $AL_{VGG}$, $AL_{ResNet}$ and $AL_{Densenet}$ denote our AL based network with base classifier of $VGG16$, $ResNet18$ and $DenseNet$, respectively. $FSL_{VGG}$ and $FSL_{Resnet}$ denote the FSL setting, where pre-trained $VGG16$ and $ResNet18$ classifiers were fine-tuned using the given data. $FSL_{CheXNet}$ denotes our implementation of CheXNet \citep{CheXNet}.

We initiate our AL experiment by taking an initial $3\%$ of the training samples, add new samples, and continue with the training till convergence. We also report results on the FSL methods by randomly selecting data from the training set and using it for training. Table~\ref{tab:AL100} summarizes the performance of the FSL methods for all $14$ pathologies when using the entire training set. Since the whole training set is used there is no scope for informative sample selection and hence AL based results are not shown. The numbers in Table~\ref{tab:AL100} are the baseline methods against which the AL based methods are compared. We also provide the percentage of diseased images having the corresponding disease label.


An important observation is for some pathologies the AUC using $AL_{DenseNet}$ is higher than obtained using $FSL_{CheXNet}$. This is especially true for those pathologies with a higher percentage representation in the dataset. This can be explained as follows. `Infiltration' has more images is higher than others. Although the images are more, the quality of the Xray images makes it difficult to distinguish multiple pathologies. As a result many images are misclassified and the corresponding AUC decreases. However, by using informative sample selection and synthetic image generation we are able to overcome these limitations and train a better classifier that performs better on the test image set. However, we do not see a corresponding AUC increase for all classes, especially those which have a small representation in the test data.

Table~\ref{tab:AL1} summarizes the AUC values for the `Infiltration' pathology class when using different percentage of training data for AL and FSL methods.  We do not show numbers for all $14$ pathologies but select `Infiltration' since it has the highest representation in the data set. Similar trends are observed for all pathologies. 

An important observation from Table~\ref{tab:AL1} is the fact that $AL_{DenseNet}$ reaches the peak performance at a fraction ($35\%$) of the entire training set. 
In this case by using $35\%$ of informative training data, $AL_{DenseNet}$ reaches an AUC similar to of $FSL_{CheXNet}$ at $100\%$ training data. With the addition of more training data the performance of all the AL methods improve. The improvement is higher initially and stabilizes after the $35\%$ threshold because it does not encounter informative samples that are significantly different from those already in the training set. 
If we set a condition that informative samples be added only if they improve the classifier AUC by atleast $0.2$ then informative samples are not added after $54\%$ samples have been used from the training set.  This observation confirms the fact that state of the art performance can be achieved with fewer training samples provided they are informative by providing qualitatively different information. Note that the $35\%$ threshold is for this dataset and will be different for other datasets. 

Table~\ref{tab:AL351} shows the performance of AL and FSL methods when using upto $35\%$ of the training data. For FSL methods the numbers are an average of $10$ runs that randomly select $35\%$ of the training data to train the classifiers. We show results only for $FSL_{CheXNet}$ since it is the best performing FSL method. We observe that the different AL based methods generate AUC values close to those of the corresponding FSL method in Table~\ref{tab:AL100} . This clearly indicates that by selecting only the informative samples, and using synthetic samples AL based methods perform as good as state-of-the-art FSL methods. 
  By using informative samples in the training step the classifier learns more discriminative information with fewer samples compared to the FSL scenario where informative and uninformative samples are used for model training. This translates to use of shallow networks and fewer computing resources.

\begin{table}[t]
\resizebox{0.5\textwidth}{!}{
\begin{tabular}{|c|c|c|c|c|}
\hline
{} & {Diseased} & \multicolumn {3}{|c|}{$100\%$ Training Samples} \\  \cline{3-5}
{} & {Image $\%$} & {$FSL_{VGG}$} & {$FSL_{ResNet}$} & {$FSL_{CheXNet}$} \\ \hline
{Atel.} & {14.24} & {0.7456} & {0.7765} & {0.8094}  \\ \hline
{Card.} & {3.42} & {0.8653} & {0.8856} & {0.9248} \\ \hline
{Eff.} & {16.43} & {0.8121} & {0.8231} & {0.8638}   \\ \hline
{Infil.} & {24.53} & {0.6923} & {0.7011} & {0.7345} \\  \hline
{Mass} & {7.094} & {0.7915} & {0.8188} & {0.8676}   \\  \hline
{Nodule} &  {7.80} & {0.7384} & {0.7532} & {0.7802} \\ \hline
{Pneu.} & {1.67} & {0.7128} & {0.7321} & {0.7680} \\ \hline
{Pneumot.} &  {6.54} & {0.8275} & {0.8412} & {0.8887} \\ \hline
{Consol.} & {5.76} & {0.7245} & {0.7422} & {0.7901} \\ \hline
{Edema}  & {2.84} & {0.8126} & {0.8347} & {0.8878} \\ \hline
{Emphy.} & {3.10} & {0.8571} & {0.8876} & {0.9371} \\ \hline
{Fibr.} & {2.08} & {0.7348} & {0.7562} & {0.8047} \\ \hline
{PT} & {4.17} & {0.7487} & {0.7634} & {0.8062} \\ \hline
{Hernia} & {0.28} & {0.8465} & {0.8694} & {0.9164} \\ \hline
\end{tabular}
}
\caption{Classification results for fully supervised learning based methods.}
\label{tab:AL100}
\end{table}

 \begin{table}[t]
\begin{tabular}{|c|c|c|c|c|}
\hline
{} & \multicolumn {4}{|c|}{Active learning ($\%$ labeled)}  \\  \hline
{} & {5\%} & {10\%}  & {15\%} & {25\%} \\ \hline 
{$AL_{VGG}$} & {0.5832} & {0.6013} & {0.6375} & {0.6691} \\ \hline
%
{$AL_{DenseNet}$} & {0.6129} & {0.6539} & {0.7053} & {0.7321}   \\ \hline
{$FSL_{VGG}$} & {0.4981} & {0.5274} & {0.5672} & {0.5952}  \\  \hline
{$FSL_{CheXNet}$} & {0.5464} & {0.5784} & {0.6091} & {0.6471} \\ \hline
\hline
{} & \multicolumn {4}{|c|}{Active learning ($\%$ labeled)}  \\  \hline
{} & {30\%} & {35\%} & {50\%} & {$75\%$}  \\ \hline 
{$AL_{VGG}$} & {0.6985} & {0.7236}  & {0.7408} & {0.7578} \\ \hline
%
{$AL_{DenseNet}$} & {0.7662} & {0.7859}  & {0.7981} & {0.8065}  \\ \hline
{$FSL_{VGG}$} & {0.6216} & {0.6462}  & {0.6782} & {0.6842}  \\  \hline
{$FSL_{CheXNet}$} & {0.6723} & {0.7047}  & {0.7142} & {0.7263} \\ \hline
\end{tabular}
\caption{AUC values for `Infiltration' using different amounts of training data for active learning and FSL methods. All results are for the test dataset}
\label{tab:AL1}
\end{table}

\begin{table}[t]
\resizebox{0.45\textwidth}{!}{
\begin{tabular}{|c|c|c|c|}
\hline
{} & \multicolumn {3}{|c|}{$35\%$ Training Samples  with AL and FSL} \\  \hline
{} & {$AL_{ResNet}$}  & {$AL_{DenseNet}$} & {$FSL_{CheXNet}$}\\ \hline
{Atel.} & {0.7701} & {0.7987} & {0.7351} \\ \hline
{Cardi.y}  & {0.8792} & {0.9208} & {0.8711}\\ \hline
{Eff.} & {0.8205} & {0.8588} & {0.8173}  \\ \hline
{Infil.} &{0.7459} & {0.7859} & {0.7047}\\  \hline
{Mass} & {0.8114} & {0.8627}  & {0.8203}  \\  \hline
{Nodule} & {0.7682} & {0.7954} & {0.7376}\\ \hline
{Pneu.} & {0.7579} & {0.7898} & {0.7341}\\ \hline
{Pneumot.} & {0.8389} & {0.8912} & {0.8412} \\ \hline
{Consol.} & {0.7387} & {0.8054} & {0.7517}\\ \hline
{Edema} & {0.8276} & {0.8829} & {0.8597}\\ \hline
{Emphy.} & {0.8831} & {0.9228} & {0.9052}\\ \hline
{Fibr.} & {0.7625} & {0.8174} & {0.7751}\\ \hline
{PT} & {0.7568} & {0.7965} & {0.7817}\\ \hline
{Hernia} & {0.8765} & {0.9187} & {0.8938}\\ \hline
\end{tabular}
}
\caption{Classification results when using $35\%$ training samples and informative sample selection for AL and FSL based methods.}
\label{tab:AL351}
\end{table}


\subsubsection{Results on the SCR Dataset}
\label{expt:scr}
We run a set of classification experiments on the SCR chest XRay database \cite{SCR} which has Xray images of $247$ patients - $93$ normal and $154$ nodule images. The images are resized to $512\times512$ pixels to ease the network's computational burden. The dataset is divided into training ($70\%$), validation ($10\%$) and test ($20\%$) folds.

In Table~\ref{tab:ALSCR} we present results for different FSL and AL based classification approaches, similar to the results on the NIH dataset in Table~\ref{tab:AL1}. The advantages of active learning are also demonstrated on the SCR dataset. This verifies the fact that our active learning approach does a good job in selecting informative samples and augmenting them across two different datasets.

In another set of experiments we used the training and validation folds of the NIH dataset and the test fold of the SCR dataset. However the classification performance decreased (AUC$<0.63$ for the best case) since images from the two datasets are very different in appearance. In principle we can use a pre-processing technique that transforms the input image to the same appearance as the training set. However the relevant method is beyond the scope of this paper. 

 \begin{table}[t]
\begin{tabular}{|c|c|c|c|c|}
\hline
{} & \multicolumn {4}{|c|}{Active learning ($\%$ labeled)}  \\  \hline
{} & {5\%} & {10\%}  & {15\%} & {25\%} \\ \hline 
{$AL_{VGG}$} & {0.5592} & {0.5841} & {0.6135} & {0.6416} \\ \hline
%
{$AL_{DenseNet}$} & {0.6022} & {0.6411} & {0.6755} & {0.7037}   \\ \hline
{$FSL_{VGG}$} & {0.4873} & {0.5147} & {0.5529} & {0.5794}  \\  \hline
{$FSL_{CheXNet}$} & {0.5380} & {0.5643} & {0.5932} & {0.6345} \\ \hline
\hline
{} & \multicolumn {4}{|c|}{Active learning ($\%$ labeled)}  \\  \hline
{} & {30\%} & {35\%} & {50\%} & {$75\%$}  \\ \hline 
{$AL_{VGG}$} & {0.6724} & {0.7013}  & {0.7184} & {0.7287} \\ \hline
%
{$AL_{DenseNet}$} & {0.7347} & {0.7518}  & {0.7754} & {0.7852}  \\ \hline
{$FSL_{VGG}$} & {0.6022} & {0.6221}  & {0.6501} & {0.6714}  \\  \hline
{$FSL_{CheXNet}$} & {0.6531} & {0.6782}  & {0.6933} & {0.7044} \\ \hline
\end{tabular}
\caption{AUC values for `Nodule' detection in the SCR dataset using different amounts of training data with active learning and FSL methods. }
\label{tab:ALSCR}
\end{table}

\subsection{Influence of Informative Sample Selection and Image Generation}

In this section, we analyze the role of informative sample selection and the subsequent generation of synthetic images. We perform classification using the following methods using a DenseNet network:

\begin{enumerate}

\item $AL_{DA}$ - uses standard data augmentation (DA) in place of CAGAN based sample generation. After identifying the most informative samples using BNN, we augment it by standard data augmentation through rotation, translation and flipping. This helps to quantify the importance of conditional GAN based synthetic sample generation.
\item $AL_{GAN}$ - uses data augmentation using a normal GAN without a conditional input variable. The GAN is trained to generate multiple images from the informative images with the same diseased label as the original. During training the input image is the original image and the outputs are rotated, translated and flipped versions of the original.
\item $AL_{DAGAN}$ - uses data augmentation using a data augmentations GANs (DAGAN) \cite{DAGAN}. The DAGAN is trained to generate multiple images from the informative images with the same diseased label as the original. 

\item $AL_{wBNN}$ - does classification without using the Bayesian Neural network for calculating informativeness. Instead it uses classification uncertainty of the image as a measure of informativeness. Classification uncertainty is obtained as the entropy of class labels.

\end{enumerate} 

%
The results are summarized in Table~\ref{tab:AL2} for `Infiltration' when using different training data percentages. Compared to our originally proposed method, $AL_{Densenet}$, $AL_{GAN}$ comes closest followed by $AL_{DA}$. This indicates that using GAN based synthetic data generation is more informative than standard DA as GAN introduces some new information compared to standard DA. Additionally, using conditional GAN based image generation generates much more informative samples. This can be explained by the fact that using CAGANs translates information from one disease class to another, whereas in standard GAN based augmentation we do not obtain the diversity information from other classes. The results also prove that use of a Bayesian neural network to determine classification uncertainty is more accurate than using classification uncertainty.



\begin{table}[h]
\begin{tabular}{|c|c|c|c|c|}
\hline
{} & \multicolumn {4}{|c|}{Active learning ($\%$ labeled)}  \\  \hline
{} & {5\%} & {10\%}  & {15\%} & {25\%} \\ \hline 
{$AL_{wBNN}$} & {0.5131} & {0.6013} & {0.6375} & {0.6691}  \\ \hline
{$AL_{DA}$} & {0.5478} & {0.5827} & {0.6237} & {0.6783}  \\ \hline
{$AL_{GAN}$} & {0.5741} & {0.6175} & {0.6653} & {0.7024} \\ \hline
{$AL_{DAGAN}$} & {0.5867} & {0.6311} & {0.6795} & {0.7191}  \\ \hline
\hline
{} & \multicolumn {4}{|c|}{Active learning ($\%$ labeled)} \\  \hline
{} & {30\%} & {35\%} & {50\%} & {$75\%$}  \\ \hline 
{$AL_{wBNN}$} & {0.6985} & {0.7236}  & {0.7408} & {0.7578} \\ \hline
{$AL_{DA}$} & {0.7014} & {0.7216}  & {0.7398} & {0.7516} \\ \hline
{$AL_{GAN}$} & {0.7253} & {0.7561} & {0.7702} & {0.7887} \\ \hline
{$AL_{DAGAN}$} & {0.7408} & {0.7732}  & {0.7812} & {0.8016} \\ \hline
\end{tabular}
\caption{AUC values for `Infiltration' using different amounts of training data with different types of data augmentation. DenseNet was the classifier network in these experiments}
\label{tab:AL2}
\end{table}

\subsection{Plausibility of Generated Images}

It is essential for generated images to be realistic and represent real medical images with disease symptoms. We test this hypothesis with the following different experiments:
\begin{enumerate}
\item We first separate the real and synthetic images into training, validation and test folds with splitting done at the patient level. This ensures that for any particular patient all the real and synthetic images are in only one of the three folds. 

\item We denote the set of real images as \textit{Real} and the set of synthetic images as \textit{Synthetic}. In one experiment we train a FSL based CheXNet classifier on real images (from its corresponding training fold) and test on synthetic images(from the corresponding test fold). This experiment is denoted as $Real-Syn$ in Table~\ref{tab:ALMix1}. In the second experiment we train a CheXNet classifier on synthetic images and test on real images. the results are summarized under $Syn-Real$. 

\item We create a new dataset by randomly mixing images from the corresponding folds of \textit{Real} and \textit{Synthetic} and denote it as \textit{Mix}. Note that in \textit{Mix} the same patient level split is maintained. We train a CheXNet on the training set and apply it on the test set. The results are summarized under $Mix-Mix$ in Table~\ref{tab:ALMix2}. 

\item We report results for other experiments such as: $1)$ $Mix-Real$ (train on \textit{Mix} and test on \textit{Real}); 2) $Mix-Syn$; 3) $Real-Mix$; 4) $Syn-Real$; and 5) $Syn-Mix$. Note that the results for $Real-Real$ are already reported in Table~\ref{tab:AL100} under $FSL_{CheXNet}$.  

\item For all the above experiments we use a FSL based CheXNet classifier.

\end{enumerate}

If the generated images are realistic then they would have similar performance as that of the base images, and mixing of the datasets should not impact performance to a large degree. Tables~\ref{tab:ALMix1} and \ref{tab:ALMix2} summarize the performance of the different experiments for the `Infiltration' disease label. 
The numbers in both tables clearly indicate that there is no significant difference in the performance of real and synthetic images. We also observe similar performance measures when mixing up the real and synthetic images. This shows that the synthetic images generated by our method are realistic and preserve the label information.

\begin{table}[h]
\begin{tabular}{|c|c|c|c|c|}
\hline
{} & {$Real$} & {$Real$}  & {$Syn$} & {$Syn$}  \\  
{} & {$-Syn$} & {$-Mix$}  & {$-Real$} & {$-Syn$} \\ \hline
{Atel.} & {0.7294} & {0.7325} & {0.7387} & {0.7314}  \\ \hline
{Cardio.} & {0.8679} & {0.8692} & {0.8754} & {0.8702}  \\ \hline
{Eff.} & {0.8104} & {0.8134} & {0.8081} & {0.8128}  \\ \hline
{Infil.} & {0.7076} & {0.7098} & {0.6976} & {0.7126} \\  \hline
{Mass} & {0.8176} & {0.8193} & {0.8247} & {0.8246}   \\  \hline
{Nodule} & {0.7418} & {0.7345} & {0.7405}  & {0.7321} \\ \hline
{Pneu.} & {0.7281} & {0.7386} & {0.7305} & {0.7397}  \\ \hline
{Pneumot.} & {0.8375} & {0.8463} & {0.8427} & {0.8359} \\ \hline
{Consol.} & {0.7532} & {0.7465} & {0.7572} & {0.7427} \\ \hline
{Edema} & {0.8642} & {0.8621} & {0.8515} & {0.8548} \\ \hline
{Emphy.} & {0.9104} & {0.9074} & {0.9135} & {0.9004} \\ \hline
{Fibr.} & {0.7729} & {0.7825} & {0.7813} & {0.7696} \\ \hline
{PT} & {0.7894} & {0.7827} & {0.7749} & {0.7792} \\ \hline
{Hernia} & {0.8874} & {0.8904} & {0.8986} & {0.9025} \\ \hline
\end{tabular}
\caption{Classification results using $FSL_{CheXNet}$ for real and synthetic images.}
\label{tab:ALMix1}
\end{table}

\begin{table}[h]
\begin{tabular}{|c|c|c|c|c|}
\hline
{} & {$Syn$} & {$Mix$} & {$Mix$} & {$Mix$}\\  
{} & {$-Mix$} & {$-Real$} & {$-Syn$} & {$Mix$} \\ \hline
{Atel.} &  {0.7263} & {0.7306} & {0.7374} & {0.7321}  \\ \hline
{Cardio.} & {0.8687} & {0.8736} & {0.8682} & {0.8741} \\ \hline
{Eff.} & {0.8192} & {0.8064} & {0.8218} & {0.8098}   \\ \hline
{Infil.} & {0.6989} & {0.7024} & {0.7114} & {0.6987} \\  \hline
{Mass} & {0.8126} & {0.8252} & {0.8171} & {0.8215}  \\  \hline
{Nodule} & {0.7461} & {0.7398} & {0.7297} & {0.7304} \\ \hline
{Pneu.} & {0.7387} & {0.7294} & {0.7254} & {0.7332} \\ \hline
{Pneumot.} & {0.8472} & {0.8433} & {0.8348} & {0.8391} \\ \hline
{Consol.} & {0.7502} & {0.7561} & {0.7481} & {0.7542} \\ \hline
{Edema} & {0.8637} & {0.8529} & {0.8674} & {0.8563} \\ \hline
{Emphy.} & {0.9028} & {0.9086} & {0.9104} & {0.9137} \\ \hline
{Fibr.} & {0.7704} & {0.7781} & {0.7834} & {0.7729} \\ \hline
{PT} & {0.7854} & {0.7803} & {0.7829} & {0.7873} \\ \hline
{Hernia} & {0.8875} & {0.8958} & {0.8982} & {0.8914} \\ \hline
\end{tabular}
\caption{Classification results using $FSL_{CheXNet}$ for real and synthetic images.}
\label{tab:ALMix2}
\end{table}

\subsubsection{Adding Synthetic Images to Training Set}

In another set of experiments we take a fixed number of real images ($700$ images in total with $50$ images from each of the $14$ classes), use them to fine tune a DenseNet and calculate the AUC values on the test set. The initial $700$ images represent approximately $1\%$ of the training fold. Subsequently, we generate synthetic images as described before and add $70$ images in each iteration ($5$ most informative images from each class). Here we do not add any of the actual training images but only the generated synthetic images. This set up is used to provide additional validation of the plausibility of generated images. 

\begin{figure}[h]
\begin{tabular}{c}
\includegraphics[height=5.9cm, width=7.99cm]{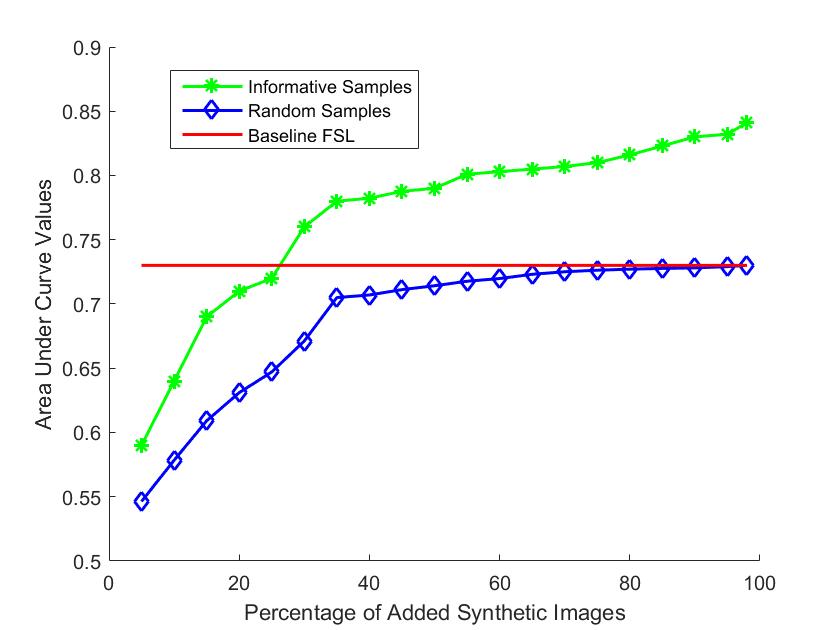} 
\end{tabular}
\caption{Change in AUC values with different number of added synthetic images.}
\label{fig:plot1}
\end{figure}

In Figure~\ref{fig:plot1} the green plot shows the change of AUC values for different number of added informative synthetic images. The axis shows the number of added synthetic images as a percentage of the number of images. We observe that as the number of synthetic images increases the AUC for `Infiltration' increases and finally surpasses the AUC of $FSL_{CheXNet}$ at the $36\%$ threshold. This is similar to our previous results where the AL based method achieves similar performance as $FSL_{CheXNet}$ (red plot) with $35\%$ of the training images.  

The blue plot in Figure~\ref{fig:plot1} shows the AUC values when we add synthetic images without choosing the most informative ones. In this case the AUC value comes close to that of $FSL_{CheXNet}$ at around $97\%$ of the training set which is observed for fully supervised learning scenario. These results clearly demonstrate that the synthetic images are as good as the real images. In a situation where we have very few real images, synthetic images generated by our method will do a good job of data augmentation and contribute to achieving state of the art performance despite fewer samples to begin with.

\subsection{Validation of Synthetic Images By Clinical Experts}

We also validate the authenticity of generated images through visual examination by clinical experts.
A $90$ minutes session was set with three experienced radiologists with over $15$ years of experience in the diagnosis of lung conditions from X-ray/CT images. The experts were tasked to perform a diagnosis on synthetically generated images that are linked (conditional part of the CAN) to a pre-determined disease/condition (e.g. infiltration, cardiomegaly, no-finding, effusion, atelectasis, nodule, emphysema). The set of images were sequentially displayed to each expert, without including any information about the condition/disease label. Each expert assessed each image and a “Yes/No” disagreement was recorded (on a different computer screen). For cases with multiple labels, a partial agreement was recorded along with the agreed condition (e.g. “Yes for effusion”).

Of the $90$ reviewed cases full agreement (i.e. exact diagnosis) was reached on  average on $61\%$ ($55/90$) cases. Partial agreement was reached on average on $72\%$ ($65/90$) cases, and on three cases the experts were not sure and mentioned that a full CT image, or lateral views would be required. For inter-expert agreement, we note that among experts an average of 71.5\% for full agreements (i.e. experts agreed on all disease labels) and of 84\% for partial agreement (experts agreed on one or more, but not all disease labels) was found. 
%

According to the experts, most of the disagreement stems from two main reasons. First, the poor resolution and sharpness of the generated images makes if difficult to get an accurate diagnosis from a single view. Poor resolution is a known issue with GANs when generating realistic images. Second, the radiologists commented several times that the generated disease/condition was very subtle and would require a sharper image to agree with the proposed label. Specifically, for nodules, the experts commented on the need of 3D information (CT image) to rule out potentially transversally imaged vessels that resemble nodules.
Regardless of the aforementioned issues with resolution and sharpness, the expert reviewers (seeing the system’s synthetically generated images for the first time) commented on the impressive ability of the system to create realistically-looking images while mimicking conditions/diseases.


\section{Conclusion}
\label{sec:concl}

We have proposed an active learning method that selectively uses new training samples featuring high levels of informativeness, as assessed by a Bayesian Neural Network. Beyond the state-of-the-art in AL, the proposed approach proposes a variant of GAN, called here CAGAN, to generate new synthetic samples of realistic-looking images while informing on class label. This feature is further used to alleviate the class imbalance problem and hence improve model's performance. 
Results on the publicly available NIH ChestXray dataset, featuring one of the largest medical image datasets, show the benefits of the proposed AL approach to yield improved accuracy and learning rates than standard data augmentation techniques and GAN-based synthetic sample generation. Our experiments demonstrate that, with about $33-35\%$ labeled samples we can achieve almost equal classification and segmentation performance as obtained when using the full dataset. This is made possible by selecting the most informative samples for training. Thus the model sees all the informative samples first, and achieves optimal performance in fewer iterations.

The performance of the proposed AL based model translates into significant savings in annotation effort and clinicians' time. These results are of great interest to the medical image computing and radiology-oriented scientific communities in need of machine learning technologies that can cope with the intrinsic difficulties of scaling up the process of curating medical images within the daily routine of medical experts. The proposed AL approach can be inserted into a Human-Machine learning system where radiology experts act as monitors and correctors of computer-generated results, which are selectively chosen for continuous model improvement, hence reducing operator's time while optimizing model learning rate. In addition, the improved learning rate of the proposed AL approach is expected to bring benefits to clinical scenarios where imaging protocols and/or vendor characteristics (e.g. future transition to 7T MRI scanners) evolve, and existing ML techniques need to feature better levels of adaptability to such changes. 

This study has some limitations and related future work that is worth mentioning. First, we note that this study focuses on the quantitative metrics evaluating the accuracy and learning rate of the model rather than assessing the realism of the synthetically generated images. This would necessitate a dedicated medical expert-review panel participating in the study to verify the class-matching of the several thousands of synthetically generated lung lesions. Second, application of the approach to other medical datasets is necessary to confirm generalization of our findings. In this regards, as future work we plan to further investigate the proposed CAGAN framework to handle multi-sequence imaging protocols (e.g. combined T1, T1-c, T2, FLAIR MRI sequences) where model uncertainty stems unequally from different image sequences. This could be used to adapt our CAGAN framework to emphasize sequence-specific patterns that need to be learned by a model, so to make it more robust to confounder effects that tend to appear more in an specific sequence. Another line of future work comprises extending the proposed approach to domain adaptability to design an active learning framework capable of adapting in a fast manner to new imaging sequences, protocols (e.g. MRI strength fields), and different imaging vendors.

\nocite{*}
\section*{References}
\bibliographystyle{model2-names}
\bibliography{CVIU_SI_AL_Ref,MyCitations_Conf,MyCitations_Journ}

\end{document}